\documentclass[runningheads]{llncs}
\usepackage[T1]{fontenc}
\usepackage{graphicx}
\usepackage{booktabs}
\usepackage[misc]{ifsym}

\usepackage{mwe}
\usepackage{amsmath}
\usepackage{mathtools}
\usepackage{xcolor,enumitem}
\usepackage{bm}
\usepackage{algorithm}
\usepackage{algpseudocode}
\usepackage{amssymb}
\usepackage{svg}
\usepackage{subcaption}   
\usepackage[hidelinks]{hyperref}

\begin{document}

\title{Counterfactual Fairness with Graph Uncertainty}


\author{Davi Valério\inst{1} \and
Chrysoula Zerva\inst{2} \and
Mariana Pinto\inst{3} \and\\
Ricardo Santos\inst{3} \and
André Carreiro\inst{3}}

\authorrunning{D. Valério et al.}

\institute{Instituto Superior Técnico, Lisbon, Portugal\and
Instituto de Telecomunicações, Lisbon, Portugal \and
Fraunhofer Portugal AICOS, Porto, Portugal  \\
\email{\{davi.giordano, chrysoula.zerva\}@tecnico.ulisboa.pt}\\
\email{\{mariana.pinto, ricardo.santos, andre.carreiro\}@aicos.fraunhofer.pt}\\}

\maketitle              
\footnotetext{Peer reviewed pre-print. Presented at the BIAS 2025 Workshop at ECML PKDD.}
\begin{abstract}
Evaluating machine learning (ML) model bias is key to building trustworthy and robust ML systems. Counterfactual Fairness (CF) audits allow the measurement of bias of ML models with a causal framework, yet their conclusions rely on a single causal graph that is rarely known with certainty in real-world scenarios. We propose CF with Graph Uncertainty (CF-GU), a bias evaluation procedure that incorporates the uncertainty of specifying a causal graph into CF. CF-GU (i) bootstraps a Causal Discovery algorithm under domain knowledge constraints to produce a bag of plausible Directed Acyclic Graphs (DAGs), (ii) quantifies graph uncertainty with the normalized Shannon entropy, and (iii) provides confidence bounds on CF metrics. Experiments on synthetic data show how contrasting domain knowledge assumptions support or refute audits of CF, while experiments on real-world data (COMPAS and Adult datasets) pinpoint well-known biases with high confidence, even when supplied with minimal domain knowledge constraints.
\end{abstract}
\section{Introduction}
\label{sec:intro}
Evaluating the bias of machine learning (ML) models is a critical step in building trustworthy, fair, and robust ML systems \cite{barocas2023fairness}. The Counterfactual Fairness (CF) criterion is an individual-level metric designed to assess bias of ML models \cite{kusnerCounterfactualFairness2017b}. According to this criterion, an ML classifier is considered fair if its prediction is the same for an original data point and a counterfactual sample, where the protected attribute (e.g., race or gender) was changed (Section \ref{sec:counterfactuals}). CF provides a more detailed perspective on discrimination rather than group fairness metrics, such as Demographic Parity \cite{zafarFairnessConstraintsMechanisms2015}, by explicitly integrating domain knowledge into Structural Causal Models (SCMs). This causal framework supports the understanding of underlying reasons for discriminatory outcomes and can identify biases that aggregate metrics may overlook \cite{kilbertusAvoidingDiscriminationCausal2017}.
To illustrate CF, consider a scenario inspired by the Dutch childcare benefits scandal \cite{favierHowBeFair2023}. Suppose a tax authority uses an ML model to predict the probability  $\hat{Y}$ that an individual has committed fraud based on their income $I \in [0,1]$, and assume we aim to evaluate whether the classifier discriminates based on nationality $N \in \{\text{local}=1, \text{immigrant}=0\}$. Under this scenario, we could hypothesize that nationality influences income and formalize this notion with the SCM:$$N \xrightarrow{} I \text{, with } I(N) \coloneq 0.5 + 0.2N+U_I  \text{ and } N\coloneq U_N$$

where $U_I$ and $U_N$ are drawn from a Gaussian and a Bernoulli distribution, respectively. In this framework, consider Bob, who is local ($N=1$), has high income ($I=0.9$), and receives a low predicted probability of fraud ($\hat{Y}=0.2$). The classifier satisfies CF if Bob’s immigrant counterfactual ($N=0, I=0.7$) also receives a similarly low $\hat{Y}$. Although this SCM has only two nodes, real-world applications may have hundreds of variables. In these cases, Causal Discovery (CD) methods can be used to find causal graphs from observational data \cite{glymourReviewCausalDiscovery2019}.

Despite its advantages, the application of CF faces one major limitation. Its results depend on a single, fully specified causal graph. However, practitioners are rarely sure of the complete graph. Instead, there may be consensus about forbidden or required causality relationships between some of the features, while many other possible connections remain uncertain. We argue that this uncertainty should be quantified and incorporated into the CF audit.

Our approach (i) leverages CD algorithms with domain knowledge restrictions, such as causal ordering, forbidden and required edges, to enumerate a bag of DAGS that respect the constraints defined by a practitioner. We then (ii) measure the entropy of the bag of DAGs to evaluate the uncertainty of the CD algorithm, and finally (iii) provide confidence bounds on the CF audit results to evaluate the uncertainty of the assessment's conclusions. Therefore, instead of auditing CF with respect to a single graph, we audit if a model is fair with respect to a set of domain knowledge restrictions and provide reliability indicators for the audit. Concretely, in this work, we:
\begin{enumerate}
    \item Propose CF with Graph Uncertainty (CF-GU), a procedure to measure and incorporate the uncertainty of graph specification with CD into CF assessments. Our approach enables the computation of dispersion metrics, informing practitioners about the reliability of a CF audit (Section \ref{sec:cfgu}).
    \item Demonstrate that our method improves the reliability of CF assessments in empirical validation using the COMPAS \cite{angwin2016compas} and Adult Census \cite{becker1996adult} datasets (Section \ref{sec:experiments}).
\end{enumerate}

\section{Background}
\subsection{Structural Causal Models}
\label{sec:scm}
An SCM $M$ formalizes our beliefs about the causal relationships between variables of a dataset \cite{pearlCausalInferenceStatistics2016}. It consists of two sets of variables $\bm{U}$ and $\bm{V}$, and a set of functional assignments $\bm{F}$ that assigns each variable in $\bm{V}$ a value based on the other variables of the model. The variables in $\bm{U}$ are called exogenous variables, and represent unmeasured noise, while the variables in $\bm{V}$ are endogenous, measured features. If we know the value of every exogenous variable, we can determine the values of every endogenous variable using the functions in $\bm{F}$. SCMs can be represented by graphical causal models. We restrict ourselves to directed acyclic graphs (DAGs), which are graphs where all the edges between nodes are directed and do not contain any cycles. In a DAG, if a variable $X$ is the child of another variable $Y$, then $Y$ is a direct cause of $X$. To illustrate, recall the fraud example from Section \ref{sec:intro}. The endogenous variables are $\bm{V} = \{N,I\}$, the exogenous are $\bm{U}=\{U_N, U_I\}$, the functional assignments are $I(N) = 0.5 + 0.2N+U_I$ and $N=U_N$ and the implied DAG is $N \xrightarrow{} I$.

Interventions can be computed from an SCM, where we alter the SCM to measure causal effects \cite{pearlCausalInferenceStatistics2016}. We intervene on a variable by ``cutting'' its incoming edges and fixing its value. For example, in an SCM $M$ with the DAG $(A \xrightarrow{} B \xrightarrow{}C \xleftarrow{}D)$, we can fix the variable $B:=b$, which yields the modified model $M_{B:=b}$ with DAG $(A\quad B \xrightarrow{}C \xleftarrow{}D)$. In this section, we have presented interventions, which will be used when estimating counterfactuals.

\subsection{Counterfactuals}
\label{sec:counterfactuals}
With an SCM, we can compute counterfactuals \cite{pearlCausalInferenceStatistics2016}. Counterfactuals are ``what-if'' statements, in which the ``if'' portion is a hypothetical condition. For example, ``what would be the value of $Y$ if $X$ had been $x_1$ instead of $x_2$'' is denoted by $Y_{x_1}(u)$, where $u$ is the noise's value when $X=x_2$. To calculate counterfactual quantities, we follow three steps: (i) abduction, (ii) action, and (iii) prediction.

In the \textbf{abduction} step, we obtain an initial condition by updating the exogenous variables $\bm{U}$ under evidence $e$. In the fraud example, our evidence was Bob's nationality and income value $(N=1, I=0.9)$. With this evidence, we calculated the exogenous variables' values that fully describe Bob (the initial state): 
$$U_N=N \text{ and } U_I=I-0.5-0.2N,$$
yielding $(U_N=0, U_I=0.2)$. In the \textbf{action step}, we intervene on some variable of interest to apply the hypothetical condition in the model. In the example, we intervened on the nationality $N:=1$. Finally, the \textbf{prediction} step calculates the value of the other variables in the model, given the initial condition and the intervention. In the example, we calculated that Bob's income if he had been an immigrant would be $I=0.7$, given that his income as a local is $I=0.9$. Note that this value would change if the causal graph or the functional assignment were different. Also, note that this formulation of counterfactuals is different from the original CF work \cite{kusnerCounterfactualFairness2017b}. While both definitions follow Pearl \cite{pearlCausalInferenceStatistics2016}, CF uses probabilistic counterfactuals \cite{kusnerCounterfactualFairness2017b}, which can also be applied when the evidence is not fully known. In this work, we compute deterministic counterfactuals in order to isolate the uncertainty due to the choice of graph for the causal model, although our procedure is extensible to probabilistic counterfactuals.

\subsection{Counterfactual Fairness}
CF uses counterfactual quantities to assess if a model discriminates individuals based on a protected attribute \cite{kusnerCounterfactualFairness2017b}. The intervention step allows for changing the sensitive feature of an individual, generating a comparable new sample. The practitioner can then test if an ML model keeps its output constant for the counterfactual instance. To define CF, let $\bm{A}$, $\bm{X}$, and $Y$ represent the protected attributes, remaining attributes, and an output of interest. Also, let $M$ be a causal model with exogenous $\bm{U}$ and endogenous $\bm{V} \equiv \bm{A} \ \cup \ \bm{X} $. A predictor $\hat{Y}$ is counterfactually fair if, for all initial values  $(x, a)$, the probability $P(\hat{Y}=y)$ is the same for the original individual and for the counterfactual $(x', a')$, where the sensitive feature was intervened upon $\bm{A}:=a'$ and the new value of $\bm{X}$  calculated with the counterfactual sampling procedure. Formally, $\hat{Y}$ is fair if
$$P(\hat{Y}_{A := a}(u)=y) = P(\hat{Y}_{A := a'}(u)=y), \ \forall \ a' \in A \textbackslash  a  ,y \in Y.$$

In this paper, we focus on ML binary classifiers and apply the CF method with the pipeline proposed in \cite{carreiro2023matrix}. Carreiro et al. proposed to count the outcome changes between instances of a dataset and their counterfactual samples, then summarize the results in a Counterfactual Confusion Matrix (CCM), from which they calculate the Positive Switch Rate (PSR) and the Negative Switch Rate (NSR). The PSR indicates the proportion of originally negative outcomes that switched to counterfactual positive predictions, while the NSR indicates the proportion of originally positive outcomes that switched to negative. Formally, for $n$ individuals, denoting as $\hat{Y}_i$ and $\hat{Y}_i^\text{cf}$ the original and counterfactual predictions for an individual $i$, respectively:

$$
\text{PSR} =
           \frac{\sum_{i=1}^{n}\mathbf{1}\!\{\hat{Y}_i=0,\;\hat{Y}^{\mathrm{cf}}_i=1\}}
                {\sum_{i=1}^{n}\mathbf{1}\!\{\hat{Y}_i=0\}}, \ \
\text{NSR} =
           \frac{\sum_{i=1}^{n}\mathbf{1}\!\{\hat{Y}_i=1,\;\hat{Y}^{\mathrm{cf}}_i=0\}}
                {\sum_{i=1}^{n}\mathbf{1}\!\{\hat{Y}_i=1\}}.
$$

\subsection{Causal Discovery}
To apply Counterfactual Fairness, we must specify a causal DAG over the variables of interest. Since the number of DAGs grows super-exponentially (e.g., $4.2\cdot10^{18}$ possible DAGs with 10 nodes), manual specification is infeasible and vulnerable to cognitive biases \cite{barocas2023fairness,oeisA003024}. In response, Causal Discovery (CD) algorithms can supply candidate graphs from observational data that may inform further debate about the causal relationships \cite{binkyteCausalDiscoveryFairness2023}. In this paper, we use Best Order Score Search (BOSS), implemented in the Tetrad Project, which searches over permutations of variables to maximize an information based score like the Bayesian Information Criterion (BIC) and yields a Complete Partially Directed Acyclic Graph (CPDAG) \cite{andrewsFastScalableAccurate2023c,scheines1998tetrad,GideonSchwarzDimensionsOfaModelBIC}. A CPDAG represents a Markov Equivalence Class (MEC) of DAGs, with directed and undirected edges, where the undirected edges mark directions that are unidentifiable from the relationships of the observational data \cite{glymourReviewCausalDiscovery2019}. To exemplify, the CPDAG $A -B-C$ represents the MEC $\{(A\xrightarrow{}B\xrightarrow{}C), (A\xleftarrow{}{}B\xleftarrow{}{}C), (A\xleftarrow[]{}{}B\xrightarrow{}C)\}$, where each DAG in the MEC imply that $A$ is independent of $C$ given $B$ $(A \perp\!\!\!\perp C \mid B)$.

To reduce the search space of CD algorithms, we can apply domain knowledge by setting a causal order \cite{scheines1998tetrad}. A causal order on a set of variables $\bm{V}$ is a strict partial order $\prec$ on $\bm{V}$. If we set $X\prec Y$, for $X,Y\in \bm{V}$, then $Y\xrightarrow{}X$ is forbidden, but $X\xrightarrow{}Y$ or `no-edge' are allowed. Equivalently, for disjoint subsets of variables $\bm{S},\bm{T} \subseteq \bm{V}$, if we set $\bm{S} \prec \bm{T}$, then every edges $t\xrightarrow{}s$, for $s\in \bm{S}$ and $t \in \bm{T}$, is forbidden \cite{pearlCausalInferenceStatistics2016}.

However, CD algorithms rely on assumptions about the dataset to ensure asymptotic correctness. For example, BOSS assumes that the data is Markov, Faithful, sufficient, i.i.d. sampled from an exponential family, and has no selection bias \cite{andrewsFastScalableAccurate2023c}. Real-world data often violates one or more of these conditions, and we argue that quantifying the graph uncertainty is important to a reliable application of learned causal graphs.

\section{Counterfactual Fairness with Graph Uncertainty}
\label{sec:cfgu}
\begin{figure}[h]
    \centering
    \includegraphics[width=\linewidth]{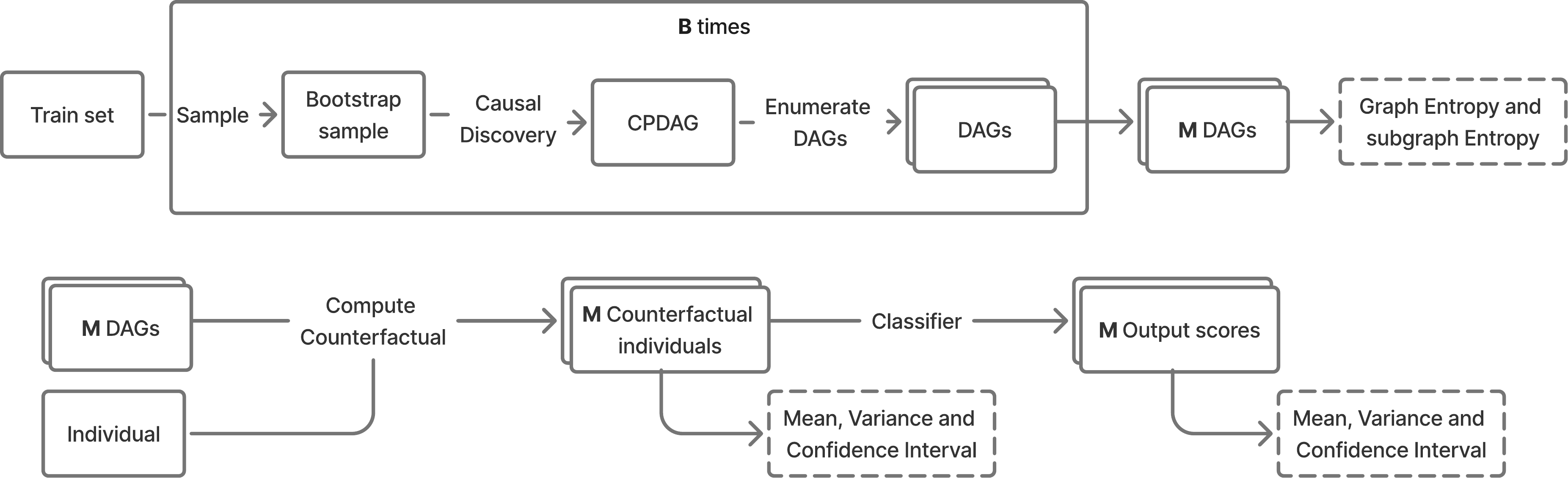}
    \caption{Counterfactual Fairness with Graph Uncertainty}
    \label{fig:cf-uncertainty}
\end{figure}

In this section, we describe and then formalize CF-GU. As illustrated in Figure \ref{fig:cf-uncertainty}, we quantify graph uncertainty in a CF assessment by bootstrapping a CD algorithm $B$ times to infer an ensemble of $M$ DAGs in total, and then by measuring the Shannon entropy. From the bag of DAGs, we derive a set of counterfactuals for a given individual and use it to evaluate CF: we feed each counterfactual into a classifier, and evaluate the variance of its output scores, as well as the variance of the PSR and NSR. A large variance indicates high uncertainty of the CF assessment due to the graph specification procedure, while a low variance indicates the opposite. 

Formally, let $(X, Y)$ be a random vector, where $X \in \mathbb{R}^d$ are observed features and $Y$ is a binary outcome. Among these features, one is a binary protected attribute denoted by $A$. We split the dataset into training and test sets, $X_{\text{train}}$ and $X_{\text{test}}$, respectively. Given a trained classifier $\phi(x)$ that outputs a score $r \in [0,1]$, we wish to assess if it is fair with respect to CF.\@ Our goal is to measure the uncertainty of the assessment when the graph selection process is driven by a CD algorithm and domain knowledge restrictions.

We start by generating $B$ bootstrap samples from $X_{\text{train}}$. For each bootstrap sample $X_{\text{train}}^b$, we apply a CD algorithm to get a CPDAG $\mathcal{C}^b$, then enumerate all DAGs in its Markov Equivalence Class. For example, if  $\mathcal{C}^b = \{\text{X1} - \text{X2}\}$, its DAGs are $\{\text{X1} \to \text{X2},\, \text{X1} \leftarrow \text{X2}\bigr\}$. We collect the DAGs from each of the $B$ bootstrap samples, and denote the final set of DAGs as $\bm{G} = \{G^m \}_{m=1}^M$. From this bag of DAGs, we compute the entropy of each edge $e$

$$H_e = -p_e\ \text{log}\ p_e - (1-p_e)\ \text{log}(1-p_e),$$

where $p_e$ is the frequency of each edge in the final set of DAGs. Finally, denoting as $\bm{E}$ the set of distinct directed edges that appeared in the $\bm{G}$, we calculate the total normalized entropy $H_{\bm{G}}$:

$$H_{\bm G} = \frac{\sum_{e} H_e}{|\bm E|\,\ln 2}.$$

A value of $H_{\bm G}$ closer to $0$ means that almost all edges are consistently present or absent, while a value closer to $1$ means that the edge probabilities are often close to $0.5$, which implies that there is higher uncertainty in the CD output.

Since we are focused on the uncertainty of a CF assessment for sensitive feature $A$, we quantify the graph uncertainty for the subgraphs formed by $A$ and its descendants. Formally, for each $G^m \in \bm G$, define the subgraph $G_A^m$, composed of $A$ and its descendants. Denoting $\bm G_A = \{G^m_A \}_{m=1}^M$, compute the total normalized entropy, as previously formalized, for the set $\bm E_A$ of distinct edges that appear in $\bm G_A$. This quantity, denoted $H_{\bm G_A}$, measures the graph confidence specifically for the descendants of $A$, which will be more important when evaluating the uncertainty of a CF assessment.

To measure CF and quantify the uncertainty of the audit, we fit an SCM $\mathcal{M}^m$ for each $G^m$ using its corresponding bootstrap training data. Then, to assess if $\phi(x)$ is CF-fair for an individual~$x_i \in X_{\text{test}}$, we generate one counterfactual with each $\mathcal{M}^m$, and feed the counterfactuals into $\phi(x)$. We denote the $M$ counterfactuals of $x_i$ as $X^i_\text{cf}$, and the output scores as $R^i_\text{cf}$. Finally, we compute the means $\bar{X}^i_\text{cf}$ and $\bar{R}^i_\text{cf}$, variances $\text{Var}(X^i_\text{cf})$ and $\text{Var}(R^i_\text{cf})$ and empirical confidence interval (CI) $\text{CI}_\alpha(X^i_\text{cf})$ and $\text{CI}_\alpha(R^i_\text{cf})$.

The final step is to propagate the uncertainty to the PSR and NSR from the operationalization of CF proposed in \cite{carreiro2023matrix}. Suppose $\hat{y}=1$ if $\phi(x)> \tau$ and $\hat{y}=0$ otherwise. Let $X_{\mathrm{cf}}^m$ be the counterfactuals generated from $X_{\mathrm{test}}$ under model $\mathcal{M}^m$. We then feed each counterfactual in $X_\text{cf}^m$ to get $\hat{Y}_\text{cf}^m$ and compute the $\text{PSR}^m$ and $\text{NSR}^m$ for the causal model $\mathcal{M}^m$. As before, we can evaluate the mean, variance, and CI of these quantities across causal models. Note that previously we assessed uncertainty for each individual, while here we quantify uncertainty in PSR and NSR across the full test set. By measuring the uncertainty of the PSR and NSR due to the graph, we know how reliable a given CF audit is. 

In summary, CF-GU incorporates the uncertainty of specifying a graph for CF audits. We start by setting domain knowledge restrictions, discovering multiple plausible DAGs, and measuring the confidence of edges with the normalized entropy. Then, by evaluating the confidence of fairness metrics (PSR/NSR), we quantify a margin of error that reveals the uncertainty of CF conclusions, given the specified domain knowledge.

\section{Experiments}
\label{sec:experiments}
\subsection{Synthetic data}
\begin{figure}[h]
    \centering
    \includegraphics[width=0.5\linewidth]{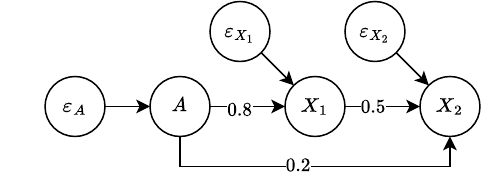}
    \caption{Ground truth DAG for synthetic experiment.}
    \label{fig:dag-synthetic}
\end{figure}

To illustrate CF-GU, we performed an experiment with synthetic data. We generated $1000$ samples with a binary sensitive feature $A$, and continuous variables $X_1$ and $X_2$ with the DAG in Figure \ref{fig:dag-synthetic}, where
$\varepsilon_A \sim \mathrm{Bernoulli}(0.5)$ and $\varepsilon_{X_{1}}, \varepsilon_{X_{2}} \sim \mathcal{N}(0,1)$. In addition, we draw a ground truth binary target $Y \sim \mathrm{Bernoulli}(p)$, where $p$ is the sigmoid $p= \frac{1}{1+e^{-z}}$ of the log-odds $z = 1.2X_1 + 0.8X_2 -0.5$.

We split the dataset into 800 training samples and 200 test samples, stratified by $Y$, and we standardized variables $X_1$ and $X_2$ to zero mean and unit variance. We then applied CF-GU, using BOSS as the CD algorithm with the degenerate Gaussian score function, a penalty of 2, and 100 bootstrap replicates, each formed by resampling the training set with replacement to produce 800 observations. We recovered graphs for $(A, X_1, X_2)$ under three different scenarios: with no domain knowledge (Low), an intermediate level of domain knowledge where we set the causal ordering $A \prec {(X_1,X_2)}$ (Medium), a high amount where we set the causal ordering $A \prec X_1 \prec X_2$ (High), ``Forbid-$X_1$'', where we removed the causal ordering restrictions, but forbid the edge $A \xrightarrow{} X_1$, and ``Forbid-$X_2$'', where we disallowed the edge $A \xrightarrow{} X_2$. Finally, we trained three classifiers using Python's scikit-learn and the library's default parameters: a Logistic Regression (LR), a Random Forest (RF) and a Gradient Boosted Classifier (GB), using $X_1$ and $X_2$ as predictors and $Y$ as the target.

With this setup, we report the results of bootstrapped CD in Table \ref{tab:synthetic_bootstrapped_cf}. Although the number of unique CPDAGs is the same across different knowledge scenarios, the total number of DAGs drops with the additional restrictions, due to a lower number of undirected edges in the CPDAGs. Similarly, with a higher amount of domain knowledge, the total normalized entropy, subgraph entropy, and the variables' variances decrease. This shows that adding the causal ordering restrictions made the CD process more confident in the output graph. When forbidding $A \xrightarrow{} X_1$, there is again higher graph uncertainty, and $\mathrm{Var}(X_2)$ is similar to the Medium setting. The entropy increases further when forbidding $A \xrightarrow{} X_2$, with the highest graph uncertainty among all knowledge scenarios. In this setting, the variances are similar to Low knowledge.

\begin{table}
\caption{Results of bootstrapped Causal Discovery for the synthetic dataset.}\label{tab:synthetic_bootstrapped_cf}
\centering
\begin{tabular}{|l|c|c|c|c|c|c|}
\hline
Knowledge & Unique CPDAGs & Total DAGs & $H_{\bm{G}}$ & $H_{\bm{G}_A}$  & $\mathrm{Var}(X_1)$ & $\mathrm{Var}(X_2)$ \\
\hline
Low    & 2 & 327 & 0.7660 & 0.6061 & 0.1462 & 0.0385 \\
Medium & 2 & 109 & 0.3672 & 0.3671 & 0.0026 & 0.0031 \\
High   & 2 & 100 & 0.1455 & 0.1455 & 0.0023 & 0.0020 \\
Forbid-$X_1$   & 2 & 209 & 0.5354 & 0.2562 & 0.0 & 0.0033 \\
Forbid-$X_2$   & 2 & 100 & 0.8328 & 0.9018 & 0.1375 & 0.0303 \\
\hline
\end{tabular}
\end{table}

\begin{figure}[h]
    \centering
    \includegraphics[width=0.7\linewidth]{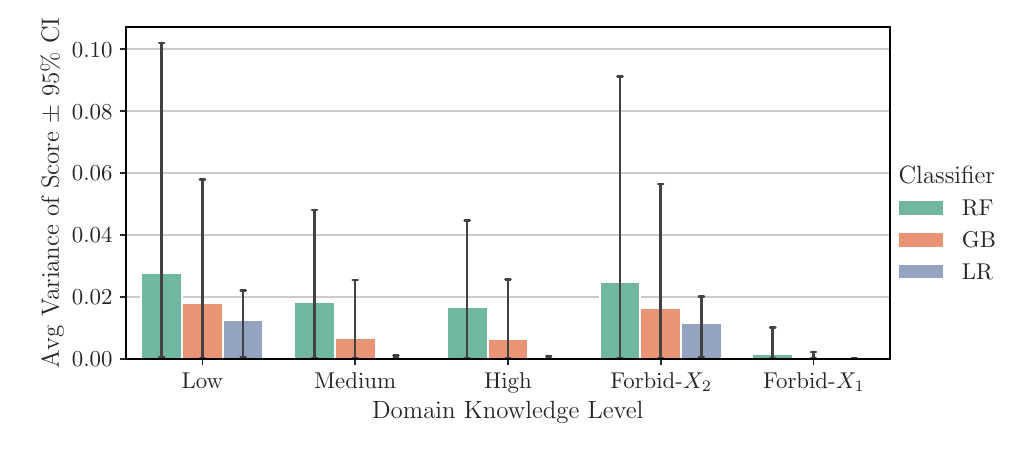}
    \caption{Average variance of score $\mathrm{Avg}_i
(\text{Var}(R^i_\text{cf}))$ (bars) and 95\% CI (whiskers) across domain knowledge scenarios and classifiers, for the \textit{synthetic} dataset.}
    \label{fig:synthetic_score_variance}
\end{figure}

We measure the variance of the score $\text{Var}(R^i_\text{cf})$ under each domain knowledge scenario for the three classifiers, then take $\mathrm{Avg}_i
(\text{Var}(R^i_\text{cf}))$ and $\mathrm{CI}_{0.05}
(\text{Var}(R^i_\text{cf}))$ across the 200 individuals. These results are shown in Figure \ref{fig:synthetic_score_variance}. In terms of $\mathrm{Avg}_i(\text{Var}(R^i_\text{cf}))$, RF is the most unstable, with values of $0.0277$, $0.0183$, and $0.0166$ for the Low, Medium, and High knowledge scenarios, respectively. GB declines from $0.0179$ to $0.0067$ and then $0.0063$, while LR is the most stable, with variance dropping from $0.0126$ to $0.0006$ and $0.0005$. The 95\% CIs of the scores are largest in the Low domain knowledge, then significantly decrease for the Medium scenario, across all the classifiers. Specifically, GB's CI drops from  $(0, 0.058)$, to $(0, 0.0255)$. For LR, the $95\%$ CI narrows from $(0, 0.022)$ (Low) to $(0, 0.001)$ (Medium), and for RF the $95\%$ CI drops from $(0.0004, 0.1021)$ (Low) to $(0.0002, 0.0481)$ (Medium). There is no significant difference in the score variance between the Medium and High settings, which indicates that the amount of domain knowledge in the Medium scenario is already enough to restrict the CD algorithm's search space and to yield a more stable graph. Under Forbid-$X_2$, the scores are similar to the Low domain knowledge, with average and $95\%$ CI $0.0248$ $(0.0003, 0.0912)$; $0.0163$ $(0.0, 0.0564)$; and $0.0115$ ($0.0004$, $0.0202$) for RF, GB, and LR, respectively. Under Forbid-$X_1$, the scores are very stable, with average and $95\%$ CI $0.0017$ $(0.0, 0.01)$; $0.0004$ $(0.0, 0.0022)$; and $0.0001$ ($0.$, $0.0001$) for RF, GB and LR, respectively. 

\begin{figure}[h]
    \centering
    \includegraphics[width=0.8\linewidth]{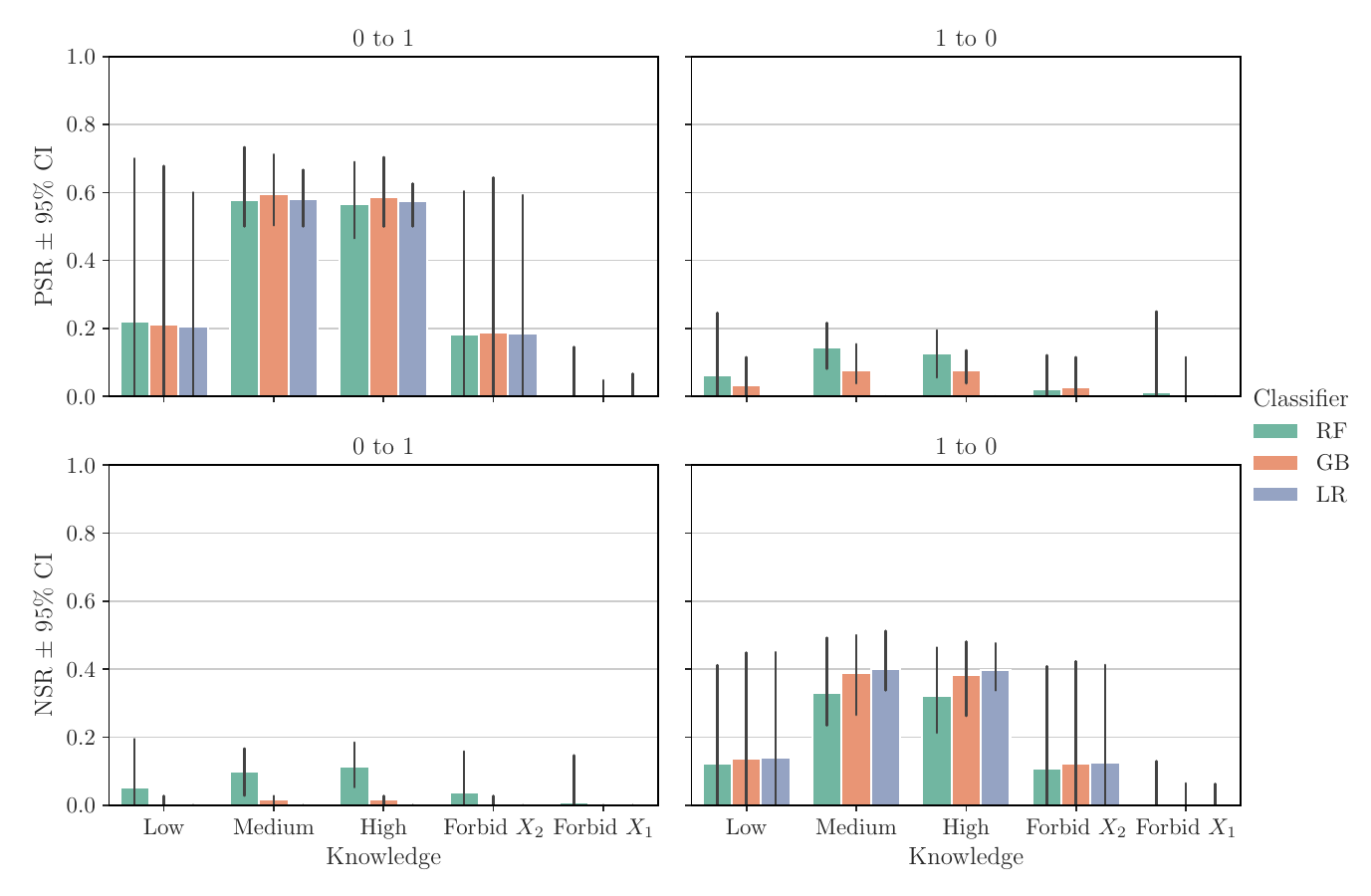}
    \caption{Average (bars) and 95\% CI (whiskers) of the PSR and NSR across causal worlds for each classifier, for the \textit{synthetic} dataset.}
    \label{fig:synthetic_fairness_metrics}
\end{figure}

Figure \ref{fig:synthetic_fairness_metrics} summarizes the PSR and NSR of the CF assessment and their graph uncertainty. Under the intervention $A: 0 \xrightarrow{} 1$, the mean PSR grows from the Low to the Medium and High knowledge scenarios, then decreases under Forbid-$X_1$ and Forbid-$X_2$. The same behavior happens for the mean NSR, under the intervention $A: 1 \xrightarrow{} 0$, across all classifiers. The confidence intervals are largest for the Low and Forbid-$X_2$ domain knowledge scenarios, and smaller for the Medium, High, and Forbid-$X_1$. Specifically, under each scenario and classifier, their specific values are shown in Table \ref{tab:synthetic_cf_metrics}

\begin{table}[ht]
\caption{PSR and NSR for GB, LR, and RF across domain knowledge scenarios for the \textit{synthetic} dataset.}
\label{tab:synthetic_cf_metrics}
\centering
\setlength{\tabcolsep}{6pt}   
\begin{tabular}{|l|l|r|r|r|}
\hline
Classifier & Metric & Low & Medium & High \\
\hline
GB & PSR & 0.210\;(0,\,0.678) & 0.595\;(0.503,\,0.712) & 0.588\;(0.500,\,0.704) \\
   & NSR & 0.006\;(0,\,0.027) & 0.017\;(0,\,0.027) & 0.017\;(0,\,0.027) \\
\hline
LR & PSR & 0.205\;(0,\,0.600) & 0.579\;(0.500,\,0.667) & 0.576\;(0.500,\,0.626) \\
   & NSR & 0.000\;(0,\,0) & 0.000\;(0,\,0) & 0.000\;(0,\,0) \\
\hline
RF & PSR & 0.221\;(0,\,0.700) & 0.577\;(0.500,\,0.733) & 0.565\;(0.466,\,0.690) \\
   & NSR & 0.051\;(0,\,0.194) & 0.098\;(0.028,\,0.167) & 0.115\;(0.053,\,0.184) \\
\hline
\end{tabular}
\medskip

\begin{tabular}{|l|l|r|r|}
\hline
Classifier & Metric & Forbid-$X_1$ & Forbid-$X_2$ \\
\hline
GB & PSR & 0.003\;(0,\,0.048) & 0.187\;(0,\,0.644) \\
   & NSR & 0.0003\;(0,\,0) & 0.006\;(0,\,0.027) \\
\hline
LR & PSR & 0.004\;(0,\,0.067) & 0.184\;(0,\,0.592) \\
   & NSR & 0.000\;(0,\,0) & 0.000\;(0,\,0) \\
\hline
RF & PSR & 0.008\;(0,\,0.145) & 0.183\;(0,\,0.603) \\
   & NSR & 0.0074\;(0,\,0.146) & 0.038\;(0,\,0.158) \\
\hline
\end{tabular}
\end{table}

In summary, the synthetic experiment shows how the score and fairness metrics of a CF audit change under different flexible domain knowledge restrictions. With increasingly more causal ordering restrictions on the graph (from Low to High), the evidence for discrimination against $A=0$ increased, and the graph uncertainty decreased. Then, by removing the causal ordering and forbidding $A \xrightarrow{} X_2$, the results indicate the absence of discrimination with significant uncertainty. Finally, by forbidding $A \xrightarrow{} X_1$, the metrics indicate the absence of discrimination with confidence. With this procedure, the practitioner can observe the impact of different domain knowledge assumptions in a practical and robust way, since it (i) only requires restrictions, instead of causal graphs, and (ii) measures the uncertainty of the results due to the assumptions. 

\subsection{COMPAS}
We evaluate CF-GU on the COMPAS dataset, a collection of court records from Broward County, Florida, used to predict the likelihood of recidivism. In this experiment, we focus on evaluating the racial bias of a classifier fitted to predict if a defendant reoffended within two years. To focus on a binary race analysis, we only keep defendants who self-identify as either African-American ($0$) or Caucasian ($1$). We then select the features: age, number of juvenile felony convictions, number of juvenile misdemeanor convictions, number of other juvenile offenses, total count of prior convictions, the self-reported sex, and the severity of the current charge, along with the race. The target variable indicates whether each defendant reoffended within two years. We explicitly acknowledge the dubious social legitimacy of this risk prediction task and the oversimplification of racial identity. Despite these issues, we use this dataset and prediction task to evaluate CF-GU in a frequently used scenario by the scientific community.

We split the data into a train set with $4920$ samples and a test set with $1230$ samples ($80\%$-$20\%$), then apply CF-GU with BOSS as the CD algorithm with the degenerate Gaussian score function, using penalty $2$ and $100$ bootstrap replicas of $100\%$ sample size. We recover graphs for the features under two different scenarios: with no domain knowledge (Low) and another scenario (High) with a causal ordering where race, age and sex are in the first tier, and the others (number of juvenile felony convictions, number of juvenile misdemeanor convictions, number of other juvenile offenses, total count of prior convictions, and the severity of the current charge) are in the second tier. We also forbid any connections between the variables in the first tier. After recovering the graph, we encode the binary categorical variables (severity of current charge and sex) into 0/1 indicators, standardize the numerical features to mean $0$ and variance $1$, then fit a linear SCM with the discovered DAGs and their respective bootstrap samples from the training data. Finally, we train three classifiers using Python's scikit-learn and default parameters: a Logistic Regression (LR), a Random Forest (RF), and a Gradient Boosted Classifier (GB), with the previously mentioned features except for the race attribute. 

With this setup, we report the results of the bootstrap CD in Table \ref{tab:compas_bootstrapped_cf}. We observe that, although adding domain knowledge decreases the total entropy (from $0.5877$ to $0.2616$), the entropy of the subdags, where the race is a root node, increases (from $0.2587$ to $0.3285$). This means that although the added domain knowledge made the CD procedure more confident in general, it became less confident about the descendants of the race node when restricted to a smaller set of variables.

\begin{table}
\caption{Results of bootstrapped Causal Discovery for the COMPAS dataset.}\label{tab:compas_bootstrapped_cf}
\centering
\begin{tabular}{|l|c|c|c|c|c|c|}
\hline
Knowledge & Unique CPDAGs & Total DAGs & $H_{\bm{G}}$ & $H_{\bm{G}_A}$ \\
\hline
Low & 57 & 1061 & 0.5877 & 0.2587 \\
High & 29 & 352 & 0.2616 & 0.3285 \\
\hline
\end{tabular}
\end{table}

\begin{figure}[h]
    \centering
    \includegraphics[width=0.7\linewidth]{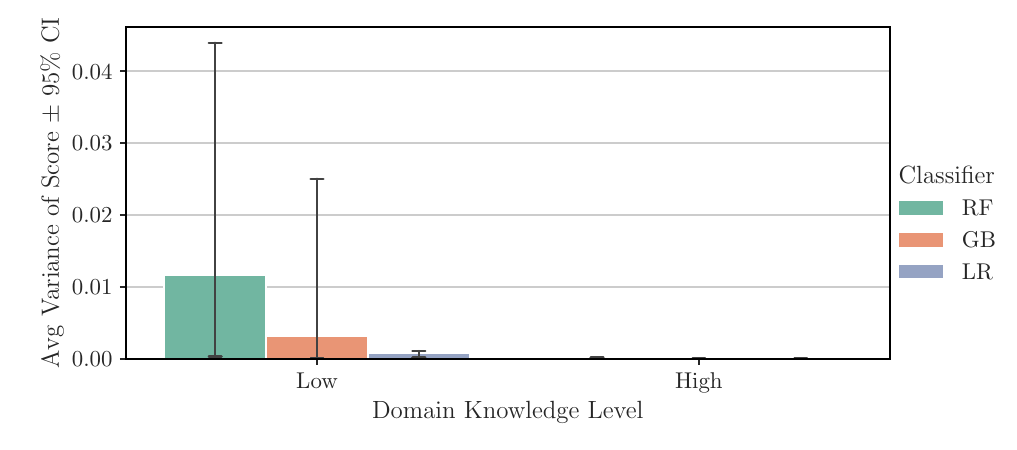 }
    \caption{Average variance of score $\mathrm{Avg}_i
(\text{Var}(R^i_\text{cf}))$ (bars) and 95\% CI (whiskers) across domain knowledge scenarios and classifiers, for the \textit{COMPAS} dataset}
    \label{fig:compas_score_variance}
\end{figure}

\begin{figure}[h]
    \centering
    \includegraphics[width=0.8\linewidth]{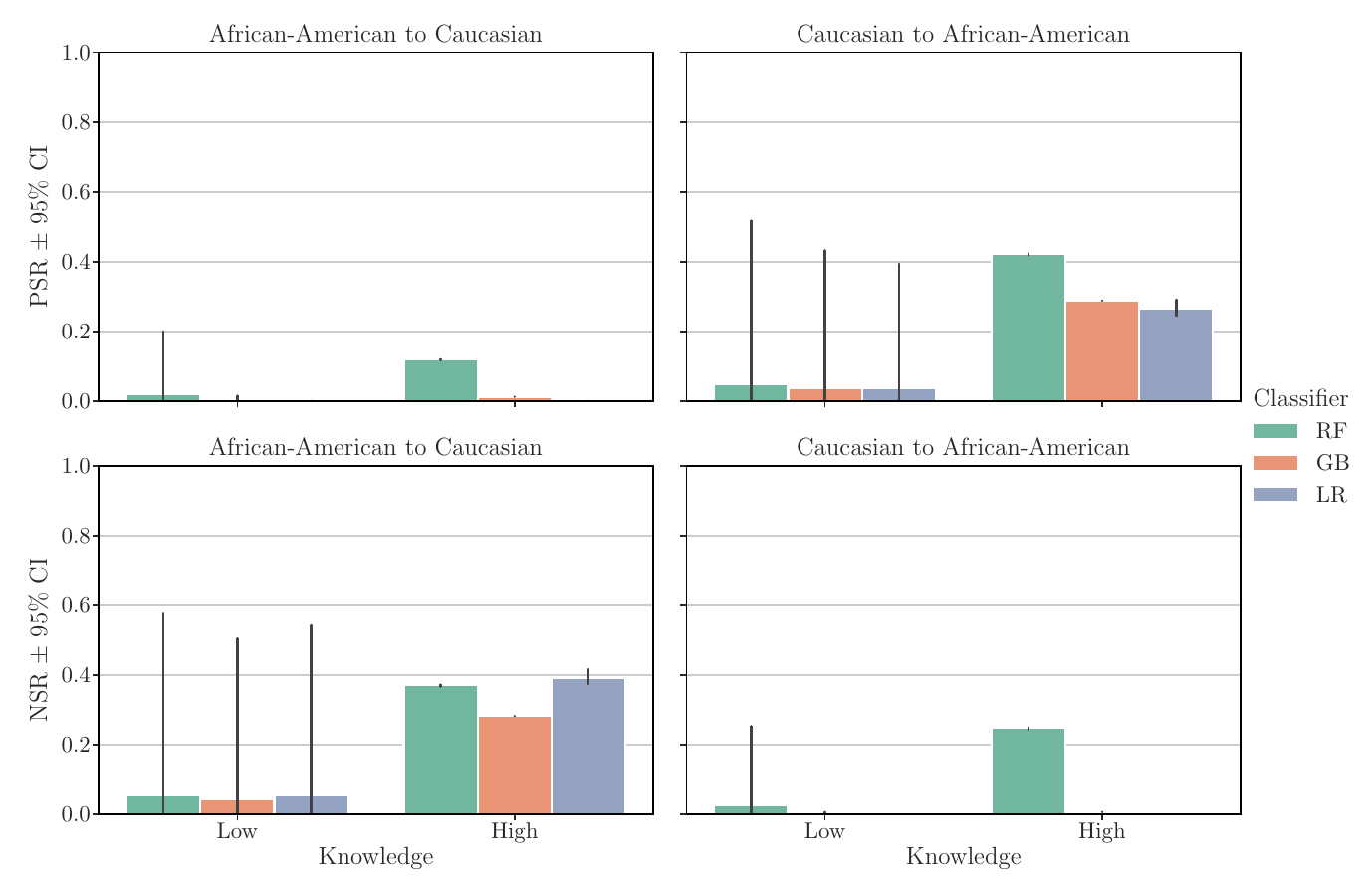}
    \caption{Average (bars) and 95\% CI (whiskers) of the PSR and NSR across causal worlds for each classifier (hue), for the \textit{COMPAS} dataset.}
    \label{fig:compas_fairness_metrics}
\end{figure}

As in the synthetic scenario, we measure the variance of the score $\text{Var}(R^i_\text{cf})$ under each domain knowledge scenario for the three classifiers, then compute $\mathrm{Avg}_i(\text{Var}(R^i_\text{cf}))$ and $\mathrm{CI}_{0.05} (\text{Var}(R^i_\text{cf}))$ across the 1230 individuals of the test set. As shown in Figure \ref{fig:compas_score_variance}, RF is again the most unstable, with an average $\mathrm{Var}(R^i_\text{cf})$ of $0.0117$ with a 95\% CI of $(0.0003,0.0439)$, followed by GB at $0.0032;(0,0.025)$ and LR at $0.0008;(0.0002,0.0011)$, all in the Low domain knowledge scenario. These variances are already small and decrease even further in the High setting. Figure \ref{fig:compas_fairness_metrics} shows the CF metrics: Under the intervention $A: \text{Caucasian} \xrightarrow{} \text{African-American}$, the average PSR grows and the CI shrinks for all classifiers, with mean and $95\%$ CI equal to 0.422 (0.417, 0.423), 0.288 (0.288, 0.288), and 0.265 (0.245, 0.291) for RF, GB and LR, respectively. Similarly, in intervention $A: \text{African-American} \xrightarrow{} \text{Caucasian}$, the average NSR increases and the CI shrinks, for all classifiers, with mean and $95\%$ CI equal to 0.372 (0.367, 0.373), 0.282 (0.282, 0.282), and 0.391 (0.374, 0.417) for RF, GB and LR, respectively.


In summary, by setting a simple restriction of causal ordering, where race, age, and sex cannot be causes of any other variable, we are able to pinpoint the bias of the classifiers with high confidence. If, instead, a practitioner postulated race was allowed to be a cause, as proposed in \cite{marcellesiRaceCause2013}, removing this restriction would allow to verify if the model remains CF-biased with respect to race.

\subsection{Adult}
We also evaluate CF-GU on the Adult dataset, which contains $48842$ census records that can be used to evaluate the bias of ML classifiers \cite{becker1996adult}. We focus on gender bias, and use the predictors age, number of years of education, hours worked per week, capital gain, capital loss, marital status, occupation, and native country to predict a binary indicator of income (if it is higher than fifty thousand). We split the data into a $39073$ train and $9769$ test sets ($80\%$-$20\%$), then apply CF-GU with BOSS as the CD algorithm with the degenerate Gaussian score function, using penalty $2$ and $10$ bootstrap replicas of $100\%$ sample size. We apply CD under two different scenarios: without any domain knowledge restrictions (Low) and another scenario with the causal ordering: (age, native country, race and sex) $\prec$ (capital gain, capital loss, education, education years, hours per week, marital status, occupation, relationship, workclass), where any connection between nodes in the first tier are also forbidden. After learning the graphs, we one-hot encode the categorical features and standardize the numerical features to mean $0$ and variance $1$. To fit a linear structural causal model on each DAG, we replace every categorical node with a set of binary indicator nodes, one per category, and replicate all of its original incoming and outgoing edges to each corresponding indicator node. We then fit a linear SCM with the encoded DAGs and their respective bootstrap samples from the training data. Finally, we train three classifiers using Python's scikit-learn and default parameters: a Logistic Regression (LR), a Random Forest (RF), and a Gradient Boosted Classifier (GB), omitting the gender attribute.

The results of bootstrap CD are in Table \ref{tab:adult_bootstrapped_cf}, where we observe an increase of both total entropy (from $0.4959$ to $0.6664$) and subgraph entropy (from $0.6506$ to $0.7634$). When restricting the causal order, we forbid the CD to produce confident but unrealistic edges. Among the plausible edges, the CD is less confident. In addition, there are new edges in the High scenario that did not appear before, as seen in the Supplementary Material. This insight is useful to guide future experimentation with restrictions of domain knowledge, where different scenarios could be tested against the confidence of the CD algorithm. If the uncertainty decreased in other plausible scenarios, it would indicate that the new restrictions are supported by the data. If, however, uncertainty increases, the data does not support the new hypothesis.

\begin{table}
\caption{Results of bootstrapped Causal Discovery for the \textit{Adult} dataset.}\label{tab:adult_bootstrapped_cf}
\centering
\begin{tabular}{|l|c|c|c|c|c|c|}
\hline
Knowledge & Unique CPDAGs & Total DAGs & $H_{\bm{G}}$ & $H_{\bm{G}_A}$ \\
\hline
Low & 8 & 470 & 0.4959 & 0.6506 \\
High & 29 & 16 & 0.6664 & 0.7634 \\
\hline
\end{tabular}
\end{table}

\begin{figure}[h]
    \centering
    \includegraphics[width=0.7\linewidth]{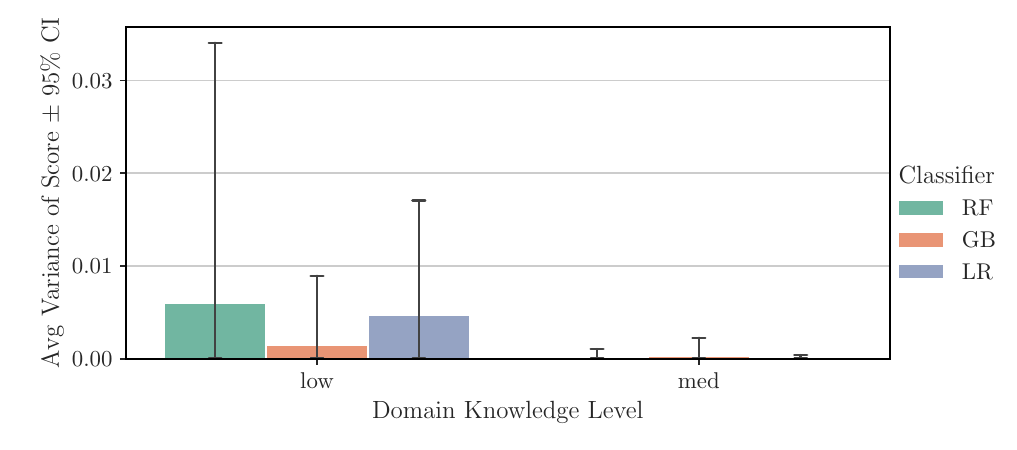 }
    \caption{Average variance of score $\mathrm{Avg}_i
(\text{Var}(R^i_\text{cf}))$ (bars) and 95\% CI (whiskers) across domain knowledge scenarios and classifiers, for the \textit{Adult} dataset}
    \label{fig:adult_score_variance}
\end{figure}

\begin{figure}[h]
    \centering
    \includegraphics[width=0.8\linewidth]{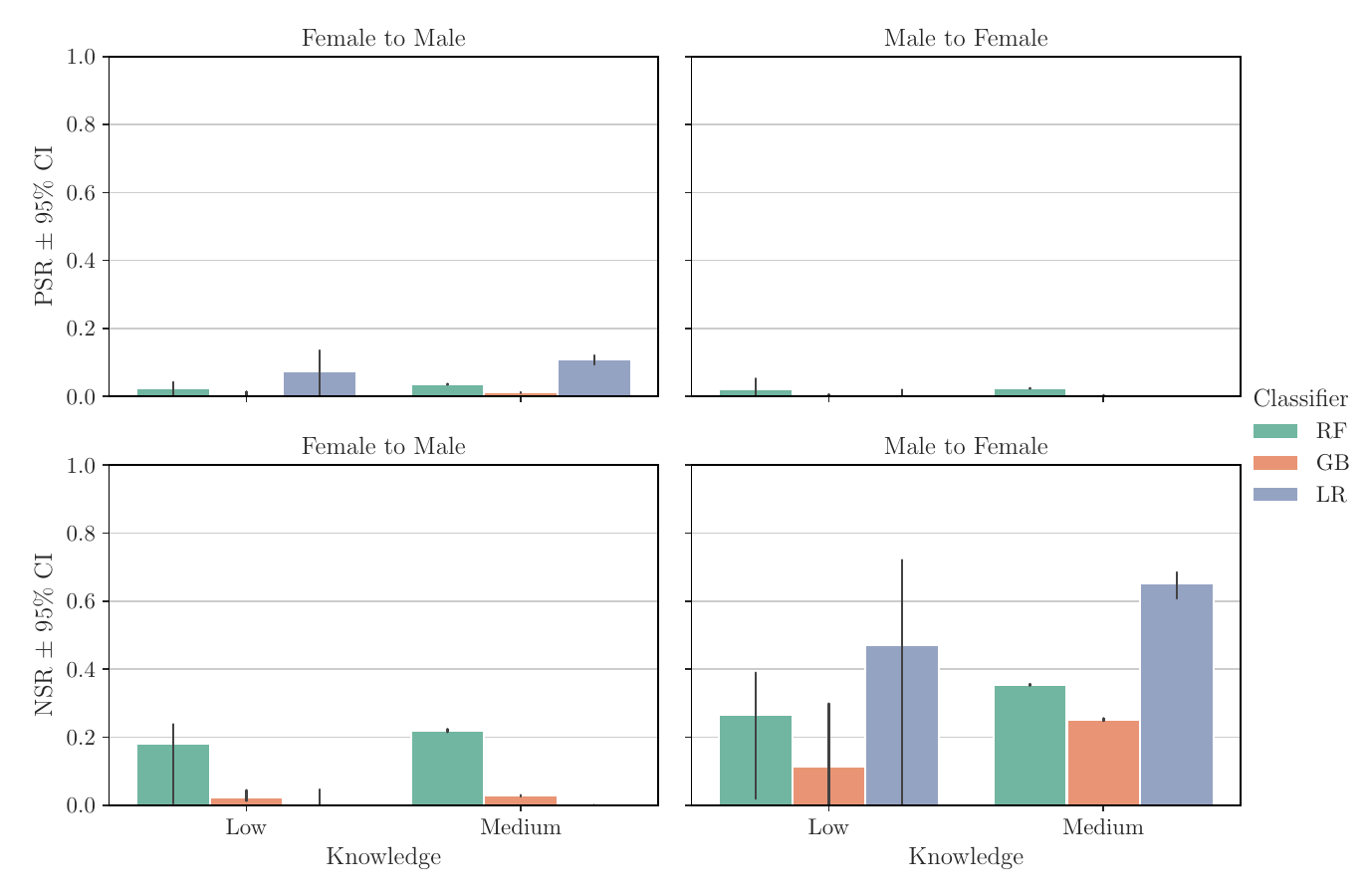}
    \caption{Average (bars) and 95\% CI (whiskers) of the PSR and NSR across causal worlds for each classifier (hue), for the \textit{Adult} dataset.}
    \label{fig:adult_fairness_metrics}
\end{figure}

Figure \ref{fig:adult_score_variance} reveals a low score variance across classifiers and domain knowledge scenarios. Under the Low setting, the average variance for RF is $0.0046$ and the $95\%$ CI $(0, 0.0232)$; for LR the average variance is $0.0028$ and the $95\%$ CI $(0, 0.0118)$; and for GB the average variance is $0.0014$ and the $95\%$ CI $(0, 0.0089)$. In the High knowledge scenario, the variances are even smaller. This indicates a high stability of the output scores for both domain knowledge scenarios. This insight is reflected in Figure \ref{fig:adult_fairness_metrics}, where the $95\%$ CI is small for both PSR and NSR under the intervention $A: \text{Female} \xrightarrow{} \text{Male}$ and for the PSR under the intervention $A: \text{Male} \xrightarrow{} \text{Female}$. The only exception is the NSR under the intervention $A: \text{Male} \xrightarrow{} \text{Female}$. The NSR is high for both RF and LR under the Low domain knowledge scenario, and the $95\%$ CIs are large for all classifiers. In the High setting, the NSR increases and the $95\%$ CI shrinks. 

In summary, the CF audit results point to the absence of discrimination with high uncertainty under the Low domain knowledge scenario. When we add domain knowledge, there is high confidence of counterfactual unfairness against Female individuals. Further experiments could attempt to determine the minimum amount of domain knowledge that is able to identify this bias in the classifiers.

\section{Related Work}
This section discusses related works. Russell et al.  recognize that different SCMs may be plausible for a given context and propose an approximation of CF that considers multiple hand-crafted SCMs simultaneously when regularizing an ML model \cite{russellWhenWorldsCollide2017}. CF-GU, differently, does not assume hand-crafted SCMs, focuses on evaluation of bias, instead of mitigation, and addresses graph uncertainty. Binkytė et al. review CD algorithms and benchmark them on synthetic and real data \cite{binkyteCausalDiscoveryFairness2023}. They focus on the sensitivity of causal fairness metrics with respect to causal graphs. However, they analyze only one DAG or CPDAG at a time and do not measure uncertainty, while CF-GU bootstraps CD to enumerate all CPDAGs compatible with the domain knowledge restrictions, then measures graph uncertainty and confidence intervals of the resulting metrics. With another perspective, Kilbertus et al. \cite{Kilbertusetal19} bound CF results with a ``counterfactual unfairness'' metric derived from unmeasured confounding effect. Their sensitivity analysis assumes that the graph is fixed and varies latent covariance, while CF-GU assumes no hidden variables and assumes the graph itself is uncertain. Finally, Zuo et al. \cite{zuoCounterfactualFairnessPartially2022} shows how to achieve CF when only a CPDAG is known. They provide a graphical test and an algorithm to select non-descendants of a sensitive attribute, then prove that predictions based on this subset are CF-fair even without a fully specified DAG. Their method aims to provide a set of features for fitting a fair predictor, given the domain knowledge restrictions, while CF-GU focuses on auditing an ML model under more flexible domain knowledge restrictions and on measuring the uncertainty of the results. 

\section{Conclusion}
An important step in building dependable, reliable, and credible ML systems is evaluating the bias of their underlying predictors. In this context, CF provides a causally informed method to evaluate the discrimination of ML models, yet CF audits are limited to a single causal graph. We introduced CF-GU, a graph uncertainty-informed approach to CF. By bootstrapping CD under domain knowledge constraints and enumerating all plausible DAGs, we quantify the uncertainty of the constraints with the Shannon entropy and provide confidence bounds for downstream fairness metrics (PSR and NSR). Experiments on synthetic data showed how different hypotheses about the causal relationships between the features yield different conclusions with varying confidence bounds in CF audits of ML classifiers. Specifically, by allowing a practitioner to flexibly define constraints for a CD algorithm, instead of fully specified causal graphs, we enable more robust and scalable CF audits. In addition, experiments on COMPAS and Adult showed that the method yields high-confidence results in line with previous work. CF-GU also opens paths for future work, such as evaluating the goodness of fit of knowledge assumptions, given the available data, and optimizing the bag of DAGs, to minimize the computational cost of fitting SCMs and generating counterfactuals for large datasets.


\begin{credits}
\subsubsection{\ackname} This work is funded by national funds through FCT - Fundação para a Ciência e a Tecnologia, I.P., with the reference 2024.07369.IACDC (DOI: 10.54499/ 2024.07369.IACDC).

\end{credits}
%
%
%
\bibliographystyle{splncs04}
\bibliography{bibliography}




\end{document}